# Modular approach to data preprocessing in ALOHA and application to a smart industry use case

Cristina Chesta
Concept BU
Santer Reply SpA
Torino, Italy
c.chesta@reply.it

Luca Rinelli
Concept BU
Santer Reply SpA
Torino, Italy
l.rinelli@reply.it

*Abstract*—Applications in the smart industry domain, such as interaction with collaborative robots using vocal commands or machine vision systems often requires the deployment of deep learning algorithms on heterogeneous low power computing platforms. The availability of software tools and frameworks to automatize different design steps can support the effective implementation of DL algorithms on embedded systems, reducing related effort and costs. One very important aspect for the acceptance of the framework, is its extensibility, i.e. the capability to accommodate different datasets and define customized preprocessing, without requiring advanced skills. The paper addresses a modular approach, integrated into the ALOHA [1] tool flow, to support the data preprocessing and transformation pipeline. This is realized through customizable plugins and allows the easy extension of the tool flow to encompass new use cases. To demonstrate the effectiveness of the approach, we present some experimental results related to a keyword spotting use case and we outline possible extensions to different use cases.

## I. INTRODUCTION

ALOHA project [1] proposed an integrated tool flow that tries to facilitate the design of DL applications and their porting on embedded heterogeneous architectures. The framework supports architecture awareness, considering the target hardware from the initial stages of the development process to the deployment, taking into account security, power efficiency and adaptivity requirements.

Reply validated the approach in the industrial environment, using the ALOHA tool flow to deploy an embedded keyword spotting system (KWS) on a IoT board (SensorTile from STMicroelectronics, based on STM32L4 ultra low-power microcontroller with Arm-Cortex M4 core) and control by voice a robotic arm (e.DO by Comau).

In detail, we evaluated the ALOHA tool flow on two possible implementations of the KWS use case: one with 6 classes and one with 13 classes.

The DL model is trained using the Google Speech Commands Dataset [2]. Both the implementations use one class to identify silence/noise and another class for unknown words, which are all the other words in the dataset that are not used in another class. This is to improve the performance of the model on an always-on board in a real scenario, where noise and other people talking are present, making it important to distinguish keywords from noise or unknown words.

The first implementation uses the 4 keywords "yes", "no", "go" and "stop" together with the 2 classes "noise" and "unknown". It is meant to control a robotic arm which is asking if it should perform an action that can be accepted with "yes" or rejected with "no", the movement corresponding to that action can be then started with "go" and stopped with "stop", even multiple times. The second implementation uses the keywords "one", "two", "three", "four", "five", "six", "seven", "forward", "backward", "go", "stop" and the 2 classes "unknown" and "noise". It is meant to control the movement of a robotic arm on its 7 axes, the selection of the axis is done with the keyword corresponding to its number and then it can be moved "backward" or "forward". The "go" and "stop" keywords can be used to start and stop the reproduction of movements from waypoints, possibly defined by voice using the previously described strategy for movements.

The realization of the two versions of KWS highlights the need of easily customize the preprocessing operations, their order, and their parameters.

## II. MODULAR ARCHITECTURE

To enable the exploitation of the tool flow in different use cases by the end users, some sections (namely: preprocessing and transforms applied to the input data, loss and accuracy metrics, split criteria for the dataset) need to be user customizable.

In previous versions of ALOHA, or in other machine learning frameworks, this kind of customization may easily require changes to the code which implements data loading and training. Code changes require a good understanding of the internal architecture and a time-consuming rebuild, which can slow down the experimentation process, introduce errors, and produce not very reusable code.

To solve these problems, a modular architecture is proposed to enable the end user to define through the GUI:

• a preprocessing pipeline, composed of modular and reusable steps, and its parameters,

• modular loss and accuracy metrics,

• and custom split criteria for the dataset.

Users can then choose among already available modules or add custom modules developed by others or themselves.

The modular architecture is composed of different types of modules. A module is composed of code and metadata describing the module's functionality, configurable parameters, and other information needed from the tool flow.

---

[1] https://www.aloha-h2020.eu/

Different types of modules are definable: transform modules implement preprocessing or data augmentation transforms; loss modules, implement loss functions; accuracy modules, implement accuracy functions; split criterion modules, define constraints on how to perform the split of the dataset in training and validation sets.

Transform modules are used to compose preprocessing pipelines. A **preprocessing pipeline** has three sections containing zero to many transform modules. Each section contains an ordered list of transform modules. The sections are executed in the given order

1. the **sample transforms** section contains a composition of transforms that is applied in parallel, with the scope of a single sample, on each sample from the dataset

2. the **dataset transforms** section contains transforms that sequentially work on the whole dataset (they have the whole dataset as scope)

3. the **batch transforms** section contains transforms that are applied by the batch iterator on each epoch for each batch given in input to the model

Each preprocessing pipeline is associated with a **preprocessing parameters file** containing a list of parameters for the transform modules used in the pipeline.

Users can define new plugins and new versions of already existing ones by uploading python scripts together with JSON metadata files that define how each specific plugin should be used in the platform. Moreover, user can select or define custom preprocessing pipelines to accommodate the data and its format for the training and testing phases through a specific user interface (see Figure 1).

The menu includes three columns representing the sections in which the transform modules are going to be applied. Each column contains, in order of application from top to bottom, cards representing the transforms. The user can intuitively drag and drop the cards representing the transforms to change the order of their execution or the section in which they are applied. From the transform card the user can access a description of the transform and input fields to configure its parameters.

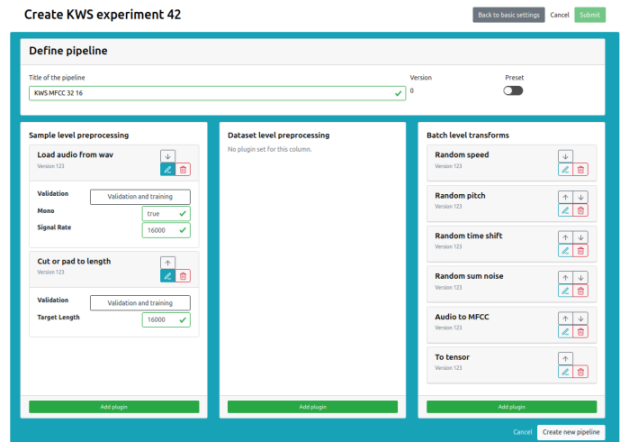

Figure 1- Configuration menu to create or edit a pipeline.

## III. EXPERIMENTS AND RESULTS

We designed a set of experiments for the KWS 6 and 13 classes implementation respectively, in order to explore multiple parameters. A total of 27 different model architectures have been trained using 15 different preprocessing pipelines, characterized by different operations and/or parameters. The plugins architecture allowed to easily compare the effect of different feature extraction methods (Mel spectrogram or Mel-Frequency Cepstral Coefficients) and of data augmentation transforms. Figure 2 shows the best results for the two different implementations, with identifier of the used pipeline and corresponding model's size. We can notice these results obtained in a semi-automatic way are comparable with state of the art in terms of accuracy [3].

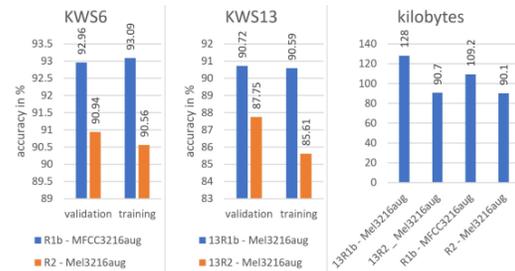

Figure 2 – Results for the KWS 6 and 13 classes and model size

## IV. CONCLUSIONS

We proposed a plugins architecture integrated into the ALOHA tool flow, which proved to be effective in supporting the exploration of different preprocessing operations and parameters in the KWS use case. Besides, allowing automatic experiment tracking and easy comparison of different configurations.

Moreover, a modular architecture also results in the extensibility of the approach to support new use cases, which is a desirable feature to make the ALOHA tool flow exploitable beyond the project's end.

In the future, we plan to further validate the methodology with extended automatic preprocessing parameters exploration.

ACKNOWLEDGMENT

This work has received funding from the European Union Horizon 2020 Research and Innovation Programme under grant agreement No. 780788. The authors would like to thank all the participants taking part in the project for their support, especially Paolo Meloni and Gianfranco Deriu from Università degli Studi di Cagliari, David Solans and Manuel Portela from Universidad Pompeu Fabra for their valuable input.